\documentclass[10pt,twocolumn,letterpaper]{article}

\usepackage{iccv}
\usepackage{times}
\usepackage{epsfig}
\usepackage{graphicx}
\usepackage{amsmath}
\usepackage{amssymb}

\usepackage{booktabs}
\usepackage{float}
\usepackage[accsupp]{axessibility}

\usepackage[pagebackref=true,breaklinks=true,letterpaper=true,colorlinks,bookmarks=false]{hyperref}

\iccvfinalcopy 

\def\iccvPaperID{8897} 

\ificcvfinal\fi

\begin{document}

\title{Time-Equivariant Contrastive Video Representation Learning}

\author{Simon Jenni \qquad Hailin Jin\\
Adobe Research\\
{\tt\small \{jenni,hljin\}@adobe.com}
}

\maketitle
\ificcvfinal\thispagestyle{empty}\fi

\begin{abstract}
We introduce a novel self-supervised contrastive learning method to learn representations from unlabelled videos.
Existing approaches ignore the specifics of input distortions, e.g., by learning invariance to temporal transformations. 
Instead, we argue that video representation should preserve video dynamics and reflect temporal manipulations of the input.
Therefore, we exploit novel constraints to build representations that are equivariant to temporal transformations and better capture video dynamics.
In our method, relative temporal transformations between augmented clips of a video are encoded in a vector and contrasted with other transformation vectors.
To support temporal equivariance learning, we additionally propose the self-supervised classification of two clips of a video into 1. overlapping 2. ordered, or 3. unordered.
Our experiments show that time-equivariant representations achieve state-of-the-art results in video retrieval and action recognition benchmarks on UCF101, HMDB51, and Diving48.
\end{abstract}
\section{Introduction}

\begin{figure}[t]
    \centering
    \includegraphics[width=\linewidth]{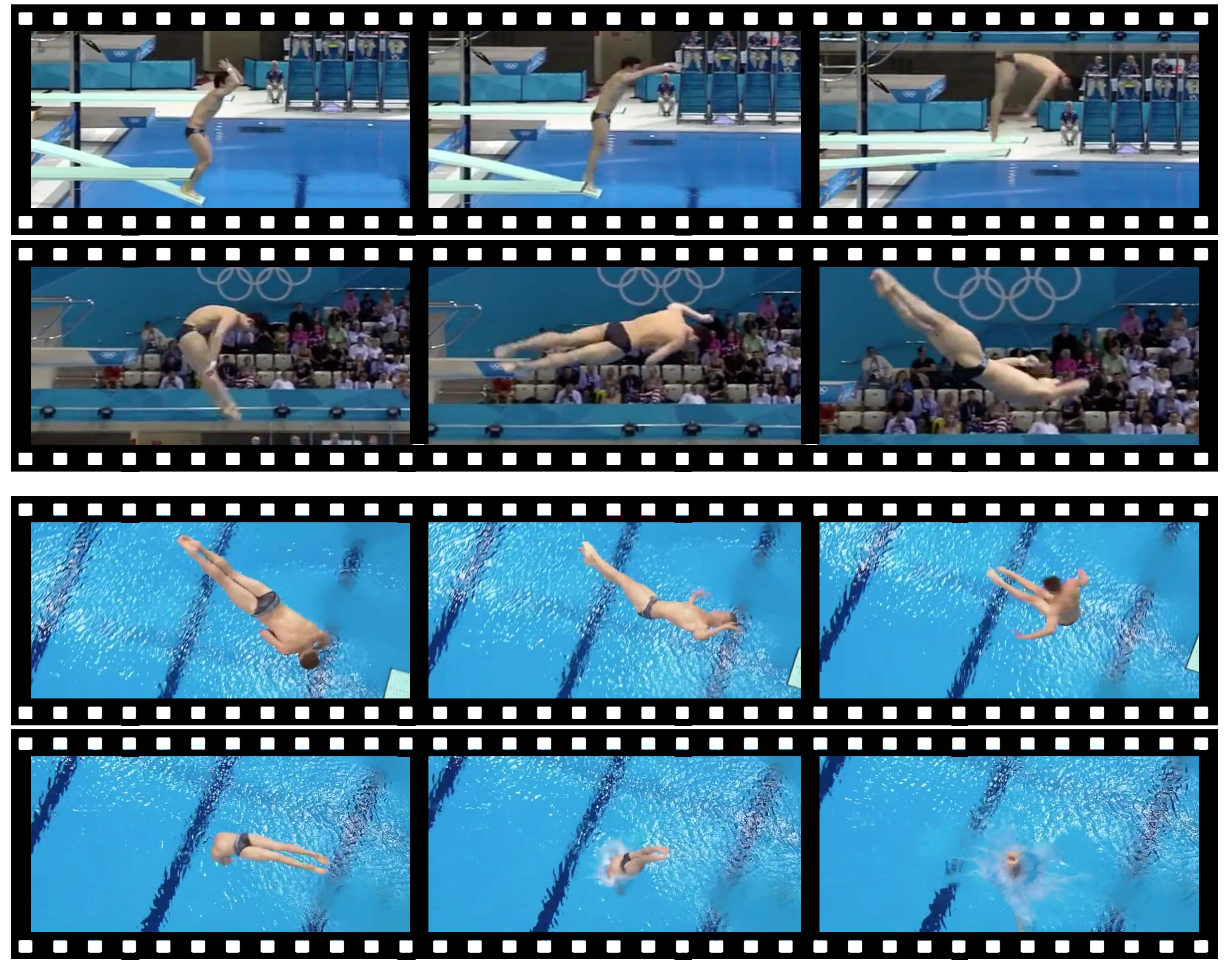}
    \caption{\textbf{The importance of motion for video understanding.} We show four clips from two videos of the Diving48 dataset \cite{li2018resound}.
    Different classes in this dataset are defined primarily by different motion patterns.  
    How should these examples be used for contrastive representation learning? 
    Besides recognizing the clips as distinct, our method exploits that the relative temporal shifts between the top two clips and the bottom two clips are identical to learn a temporally equivariant representation. 
    This promotes detailed learning of motion patterns. 
    }
    \label{fig:diving}
\end{figure}

A general video representation should accurately capture both scene appearance and dynamics to perform well on various video understanding tasks, \eg, action recognition or video retrieval. 
While large annotated video datasets \cite{zisserman2017kinetics,abu2016youtube} advanced the state-of-the-art in video understanding, such annotations are costly.
Furthermore, supervised learning from sparse action labels can introduce biases towards appearance and neglect features related to motion when actions are recognizable from static frames \cite{li2018resound,schindler2008action}.
\begin{figure*}[t]
    \centering
    \includegraphics[width=0.9\linewidth]{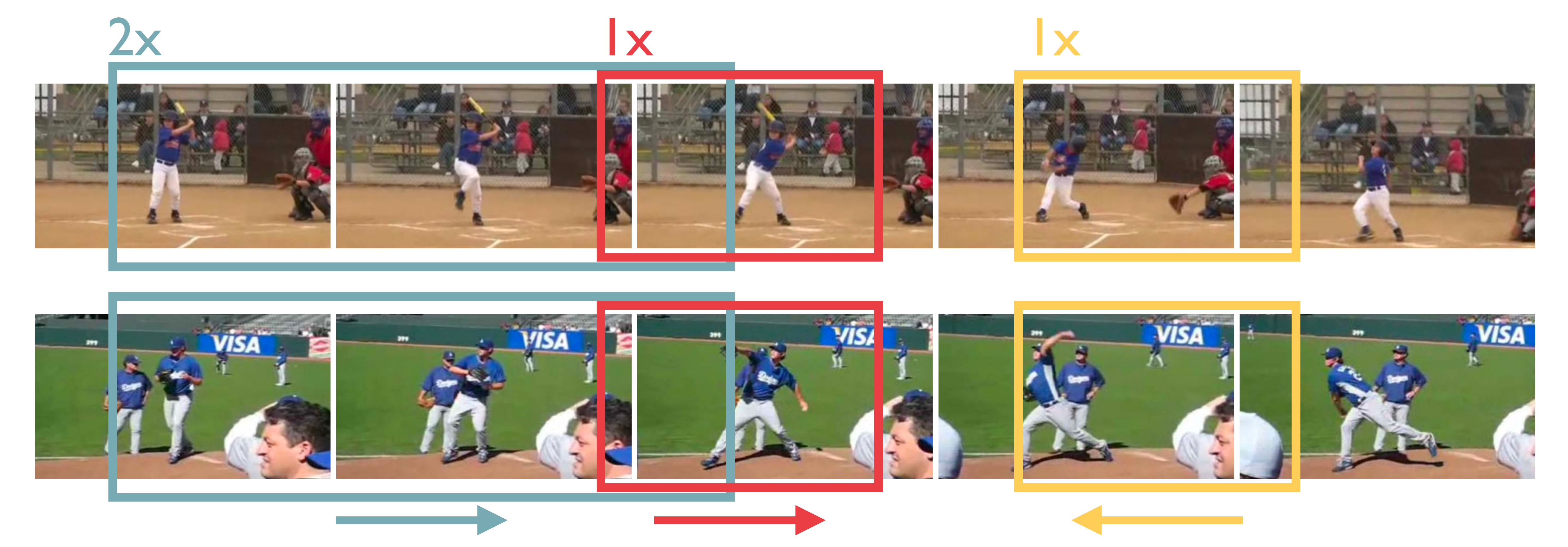}
    \caption{ \textbf{Illustration of temporal transformations and the proposed learning task.} The colored rectangles illustrate three different temporally augmented clips extracted from two videos. The playback speed is indicated above the box ($2\times$ speed resulting in double the temporal extend with the same clip size) and the playback direction with an arrow below the box. The goal of our contrastive equivariance model is to recognize identical relative transformations, \eg, recognize that the temporal transformations between blue and yellow clips in both videos are identical. 
    To support the equivariance learning, we introduce a new auxiliary task of classifying two clips into 1. non-overlapping with correct order (blue \& yellow), 2. overlapping (blue \& red), 3. non-overlapping with incorrect order (yellow \& red).  }
    \label{fig:transforms}
\end{figure*}
We consider self-supervised learning (SSL) \cite{Carl2015} as a possible solution, enabling both the training on abundant unlabelled videos and biasing learning towards video dynamics.

In SSL, learning tasks are created for which supervision does not require human labor.
On images, state-of-the-art SSL approaches based on contrastive learning are approaching or exceeding supervised pre-training.
Fundamentally, the task in these methods is to discriminate different training examples while simultaneously learning invariance to a set of input transformations, \eg, of the spatial domain via random cropping and horizontal flipping.
How should the additional temporal domain in videos be used for SSL? 

A straightforward approach would be to learn invariance to temporal input transformations.
Since such invariance could encourage the representation to disregard video dynamics, prior works proposed learning distinctiveness instead \cite{patrick2020multi,dave2021tclr}, \ie, treating different temporal crops from the same video as  distinct examples.
In contrast, rather than merely learning to recognize two temporal augmentations as different, our main idea is to learn precisely \emph{how} they differ.
We, therefore, propose to exploit additional free supervision in the form of the known temporal transformations. 
Besides learning that two clips show different moments in time, the model will also learn the clips' temporal order and even the temporal shift between them.  
This should improve the learning of dynamics and increase the models' temporal reasoning capabilities, which is crucial in situations where motion is the primary discriminating feature (\eg, see Figure \ref{fig:diving}).  
The recognition of input transformations requires that the model represents input changes in a predictable way, \eg, when the learned representation is equivariant to the transformation.

In this paper, we propose a contrastive approach for learning representations that exhibit such equivariance to temporal input transformations. 
In our method, we encode the relative temporal transformation between two input clips in a feature-vector and contrast it with other, distinct, relative transformation vectors in the mini-batch.
Therefore, a positive pair for contrastive equivariance learning results from applying the same relative transformation to two different examples.
This framework is very flexible, as it allows us to encode the desired equivariance in a set of input transformations.
Aside from standard video augmentations (\eg, random spatial and temporal cropping), we also explore equivariance to reverse playback and playback at higher speeds. 

It turns out that learning temporal equivariance is considerably more difficult than learning spatial equivariance (which we also study in experiments). 
However, temporal equivariance and the resulting preservation of motion features lead to much-improved transfer performance. 
To increase the network's sensitivity to motion patterns and improve the temporal equivariance learning, we propose as a novel pretext task the three-way classification of two clips into 1. non-overlapping with correct temporal order, 2. overlapping, 3. non-overlapping with an incorrect order. 
Additionally, we also pose as auxiliary tasks the classification of the playback speed and playback direction as proposed in prior works \cite{wei2018learning,benaim2020speednet}.
All these auxiliary tasks align with the temporal equivariance objective, the optimization of which they support. 
Figure \ref{fig:transforms} illustrates the considered transformations and the proposed learning tasks. 


\noindent \textbf{Contributions.} To summarize, we make the following contributions: 1) We introduce a novel contrastive learning approach to learn representations equivariant to a set of input transformations; 2) We study how equivariance to temporal or spatial transformations affects the quality of the learned features; 3) We introduce the pretext task of clip overlap/order prediction to support the temporal equivariance learning; 4) We show that time-equivariant representations achieve state-of-the-art transfer learning performance in several action recognition and video retrieval benchmarks on UCF101 \cite{soomro2012ucf101}, HMDB51 \cite{hmdb51}, and Diving48 \cite{li2018resound}.

\section{Related Work}

\noindent\textbf{Contrastive Learning Methods.}
Current state-of-the-art methods in unsupervised image representation learning are primarily based on contrastive learning. 
The origin of these methods can be traced to instance discrimination tasks \cite{dosovitskiy2015discriminative}. 
Wu \etal \cite{wu2018unsupervised} proposed a non-parametric formulation of this task based on a noise-contrastive estimation that allowed scaling instance discrimination to a large number of instances.
\cite{chen2020simple} proposed several improvements regarding architecture designs, stronger augmentations, and instance discrimination among large mini-batches.
Some methods rely on a queue of past negatives of momentum-encoded examples \cite{he2020momentum} to lessen the need for large batches.
Other methods remove the need for explicit negatives altogether \cite{grill2020bootstrap,chen2020exploring}.
Several recent works proposed contrastive approaches beyond instance-level discrimination, \ie, they learn a grouping or clustering of examples using the contrastive framework \cite{caron2020unsupervised,wang2020unsupervised}.
Fundamental to these approaches is the learning of transformation invariance. \cite{xiao2020should} showed that it could be beneficial also to learn distinctiveness to image transformations depending on the downstream tasks.
In contrast, we propose a contrastive approach to learn equivariance to a set of input transformations and demonstrated its benefits on video representation learning.

\noindent\textbf{Contrastive Video Representation Learning.}
Several works have explored the use of contrastive learning on videos. 
Some propose an extension to videos by adding spatially consistent temporal cropping to the set of augmentations \cite{qian2020spatiotemporal}.
Others propose to treat temporally augmented clips as distinct instances \cite{dave2021tclr,patrick2020multi}. 
Some works explored the combination of contrastive learning with other SSL tasks via multi-task learning \cite{bai2020can,wang2020self,tao2020self}.
Multi-modal contrastive learning on video has also shown promising results. \cite{han2020self} propose a joint contrastive training between RGB and optical flow. Other works rely on weak supervision in the form of text \cite{miech2020end} or exploit the audio accompanying the video \cite{afouras2020self,alwassel2019self}.
Related methods extract different "views" of the data from different intermediate layers of the network \cite{devon2020representation,xue2020self}.
Our method focuses only on the raw RGB data and goes beyond temporal distinctiveness by learning equivariance. SSL pretext tasks support this equivariance learning. 

\noindent \textbf{Temporal Self-Supervision.}
Self-supervised learning comprises pretext tasks where supervision does not involve human labor. 
This approach proved effective on images where pretext task include predicting the location of image patches \cite{Carl2015,noroozi2016unsupervised}, predicting color from grayscale images \cite{zhang2016colorful,zhang2016split,larsson2017colorproxy}, inpainting image patches \cite{pathakCVPR16context}, clustering \cite{caron2018deep,zhuang2019local}, or predicting image transformations \cite{gidaris2018unsupervised,jenni2018artifacts,jenni2020steering}.
Several methods exploit video for self-supervised image representation learning, \eg, via tracking \cite{wang2015unsupervised,owens2016ambient,gordon2020watching}.
Besides adapting image SSL tasks to videos \cite{jing2018self,han2019video,sun2019contrastive,devlin2018bert}, several works proposed using temporal self-supervision in videos, \eg, in the arrangement of video frames \cite{misra2016shuffle,brattoli2017lstm,fernando2017self,lee2017unsupervised,lee2017unsupervised}, the temporal arrangement of video clips \cite{xu2019self,kim2019self}, the arrow of time \cite{wei2018learning} or the video playback speed \cite{epstein2020oops,benaim2020speednet,yao2020video,jenni2020video}.
Our method leverages such SSL tasks to guide and improve temporal equivariance learning. 

\noindent \textbf{Learning Equivariant Representations.}
Translational equivariance is one of the defining features of CNN architectures \cite{lenc2015understanding}.
Several works have proposed network designs that endow models with additional equivariances, \eg,  capsule networks \cite{hinton2011transforming,hinton2018matrix}, group equivariant convolutional networks \cite{cohen2016group}, or harmonic networks \cite{worrall2017harmonic}.
Equivariance has also been explored for representation learning on images, \eg by considering a discretized transformation space and predicting multiples of $90^{\circ}$ rotations \cite{gidaris2018unsupervised}, or regressing the parameters of a relative transformation between two input images \cite{zhang2019aet}.
Ours is a more general learning approach that is not limited to discrete or parametrized transformations.

    

\section{Model}

\begin{figure*}
    \centering
    \includegraphics[width=0.9\linewidth]{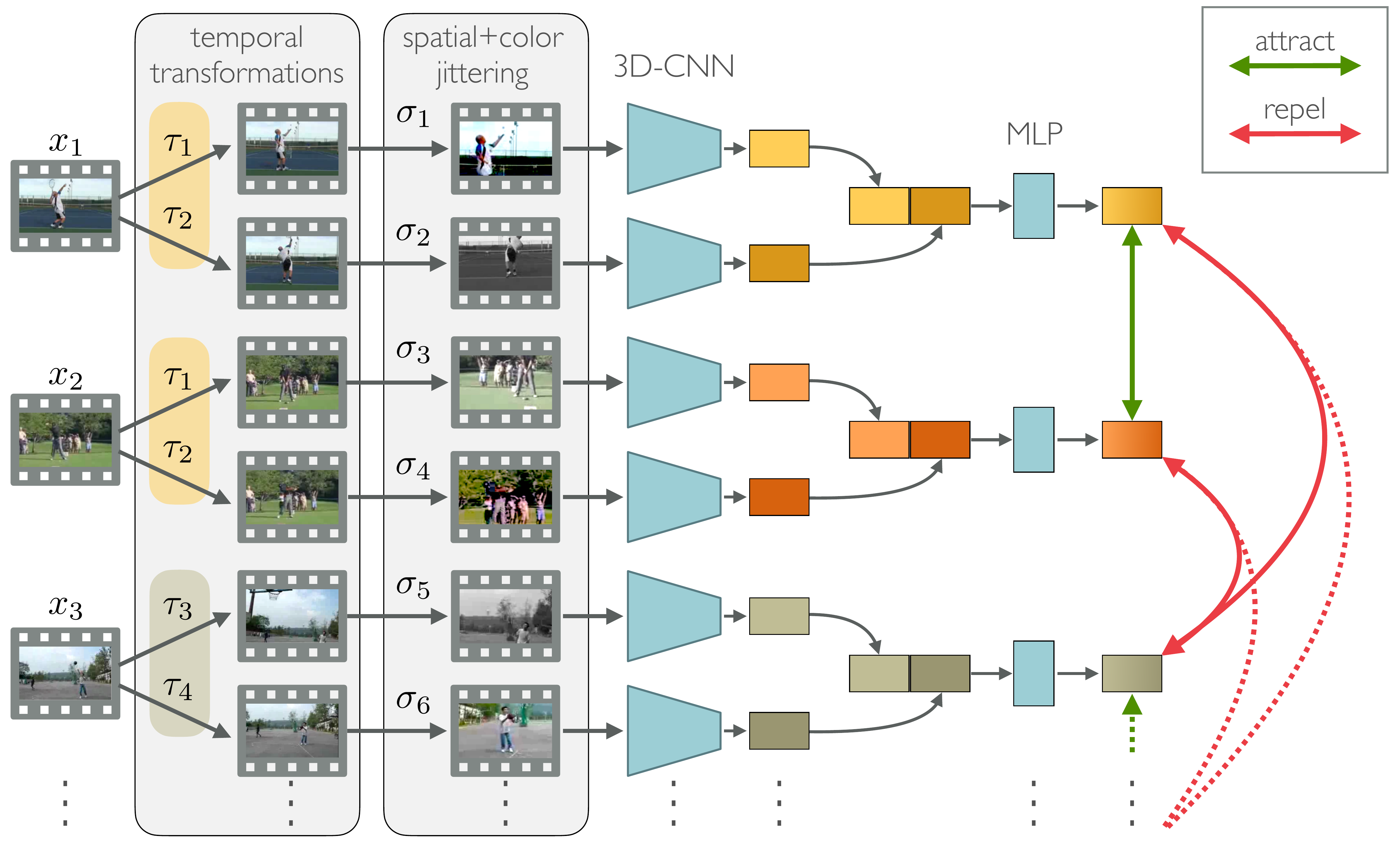}
    \caption{\textbf{Overview of the time-equivariant contrastive learning model.} On the left, we show example training videos that are passed through two sets of input transformations. First, we apply temporal transformations (\eg, temporal cropping), which define the set of desired equivariances. We apply the same temporal transformations to two training examples in the batch to enforce the equivariance constraint. Samples then pass through further spatial and color augmentations, which represent desired invariances. All the examples are then encoded in the feature space of some shared 3D-CNN. The feature vectors of the two augmentations for a given video are then concatenated and fed through an MLP that encodes the relative temporal transformation. These vectors are then attracted to the vectors encoding the same transformation and contrasted with other vectors in the batch representing distinct transformations. }
    \label{fig:equi_model}
\end{figure*}

Our approach combines a temporal equivariance objective (Sec.~\ref{sec:equi} and Fig.~\ref{fig:equi_model}) with an instance discrimination objective (Sec.~\ref{sec:inst} and Fig.\ref{fig:inst_model}). A set of auxiliary SSL tasks which are aligned with the equivariance objective, guide the network towards learning temporal features (Sec~\ref{sec:aux}). 

Let $\mathcal{D}=\left\{x_{1}, x_{2}, \ldots, x_{N}\right\}$ be a set of unlabelled training videos. 
Further, let $F$ denote the neural network we want to learn and let $F(x_i) \in \mathbb{R}^D$ be the learned representation of $x_i$.
Our goal is to learn network weights such that $F(x_i)$ is equivariant to a set of temporal transformations $\mathcal{T}$ and invariant to another set $\mathcal{S}$ consisting of spatial augmentations and color jittering. 
Concretely, for $\sigma \sim \mathcal{S}$ we desire $\forall x\in \mathcal{D}: F_\phi(\sigma(x)) \approx F(x)$ and for $\tau \sim \mathcal{T}$ we want $\forall x\in \mathcal{D}: F_\phi(\tau(x)) \approx \Bar{\tau}(F(x)) $. 
The function $\Bar{\tau}: \mathbb{R}^D \mapsto \mathbb{R}^D$ represents a transformation in feature space that corresponds to the input transformation $\tau$.
We can find an example of equivariance in the feature maps of CNNs: Spatial shifts in the input are reflected in corresponding shifts in the feature map.  
In contrast, we do not fix the behavior of $\Bar{\tau}$.
The key property we are interested in is that each $\tau \sim \mathcal{T}$ is recognizable from the pair $\big(F(x), F(\tau(x))\big)$ irrespective of the example $x \in \mathcal{D}$.

\subsection{Temporal Equivariance Learning}\label{sec:equi}
Different approaches exist to obtain equivariance to a set of transformations $\mathcal{T}$, depending on the nature of $\mathcal{T}$. 
Before introducing our approach, we will discuss some existing examples from the literature. 

When the set is finite, \ie, $\mathcal{T}=\{\tau_1, \ldots, \tau_k\}$ and when it is also possible to identify $\tau_i$ from a single example $\tau_i(x)$, then standard $k$-way classification is an option. 
Examples of this setting can be found in the classification of playback direction \cite{wei2018learning} and the recognition of playback speeds \cite{epstein2020oops,benaim2020speednet,yao2020video} or other distinct temporal transformations \cite{jenni2020video,misra2016shuffle}.

If the set $\mathcal{T}$ is finite but the $\tau_i$ are not identifiable from $\tau_i(x)$ alone, then we have to consider the pair $\big(x, \tau_i(x)\big)$ to classify the relative transformation between $x$ and $\tau_i(x)$. 
Tasks concerning the temporal ordering of clips \cite{xu2019self,kim2019self} somewhat resemble this case (although they typically consider more than two clips).

The more general case is when the set $\mathcal{T}$ is infinite. 
If all $\tau_i \in \mathcal{T}$ are parametrizable, \ie, $\tau_i$ can be described by some parameters $\theta \in \mathbb{R}^d$, then the regression of $\theta$ from $\big(x, \tau_i(x)\big)$ can be used to learn equivariance. 
An example of this can be found on images \cite{zhang2019aet} where $\tau_i$ are projective transformations.
Even when the parametrization of complex relative transformations is possible, this approach is not practical: Various classification tasks for discrete parameters (\eg, playback direction) have to be balanced with regression tasks for continuous parameters. 

We aim to learn equivariance in an even more general case where we can access data containing at minimum two pairs with identical relative transformations. 
That is, the data contains pairs $\big(\tau_p(x_i), \tau_q(x_i)\big)$ and $\big(\tau_p(x_j), \tau_q(x_j)\big)$.
This approach does not require a parametrization and also includes the case of non-parametric transformations.

\noindent\textbf{Equivariance via Transformation Discrimination.}
In what follows, we will omit the application of $\sigma \sim \mathcal{S}$ to each example $x_i$ to keep the notation uncluttered. A different $\sigma$ is applied implicitly to each occurrence of an example $x_i$, therefore encouraging invariance to $\mathcal{S}$ in the process.
We represent the relative transformation between $\tau_p(x_i)$ and $\tau_q(x_i)$ by $\psi^{pq}_i  = \psi \big([F(\tau_p(x_i))^\top,F(\tau_q(x_i))^\top] \big) \in \mathbb{R}^D$, where $\psi(\cdot)$ denotes a multi-layer perceptron, taking as input the concatenation of the feature vectors of $\tau_p(x_i)$ and $\tau_q(x_i)$. 
We want $\psi^{pq}_i$ to be similar to $\psi^{pq}_j$ for $i\neq j$, and dissimilar to $\psi^{rs}_j$ for $p,q\neq r,s$.
This can be achieved using the contrastive learning framework, by otpimizing
\begin{equation}
    \mathcal{L}_{\operatorname{equi}}=-\mathbb{E}\left[\log \frac{d(\psi^{pq}_i , \psi^{pq}_j)}{d(\psi^{pq}_i, \psi^{pq}_j) +\sum_{rs\neq pq} d(\psi^{pq}_i , \psi^{rs}_k)}\right],
    \label{eq:equi}
\end{equation}
where 
\begin{equation}
    d(\mathbf{x},\mathbf{y}):= \exp \left(\frac{1}{\lambda}  \frac{ \mathbf{x} \cdot \texttt{stopgrad}(\mathbf{y})}{\Vert \mathbf{x} \Vert_2 \Vert \texttt{stopgrad}(\mathbf{y}) \Vert_2} \right)
    \label{eq:sim}
\end{equation}
is a measure of similarity between two feature vectors, with $\lambda = 0.1$ being a temperature parameter and $\texttt{stopgrad}(\cdot)$ indicating that gradients are not back-propagated through the argument (this is similar to \cite{chen2020exploring}).
Note that, in practice, the summation in the denominator of Eq.\ref{eq:equi} is performed over other examples in the same training mini-batch.
The training on the transformation discrimination objective is illustrated in Figure \ref{fig:equi_model}.

\subsection{Auxillary Temporal SSL Objectives}\label{sec:aux}

In practice, we find that learning equivariance to temporal transformations $\mathcal{T}$ by optimizing Eq.\ref{eq:equi} alone is difficult.
Initial network parameters do not seem to capture temporal features well, and the optimization stays stuck in a bad region of the parameter space. 
To alleviate this issue, we exploit SSL tasks that use auxiliary signals that come for free with the transformations $\tau$ and are aligned with the equivariance objective.
The purpose of these tasks is to steer the network $F$ towards capturing temporal features related to video dynamics. 
This, in turn, helps the network in optimizing the equivariance objective in Eq.\ref{eq:equi}.
We explore three types of auxiliary SSL tasks, 1. classification of the playback-speed, 2. classification of the playback direction, and 3. classification of the overlap or order of the two compared clips. We describe each of these tasks in detail below.\\

\noindent\textbf{Speed Classification.} 
In this case, the temporal transformations contain changes in the playback speed. 
We consider up to four different speed classes corresponding to a $1\times$, $2\times$, $4\times$, and $8\times$ increase of the original playback speed. 
We train a separate non-linear classifier on the feature representation $F(x)$ to classify these speed types.
This task has already been proposed and explored in several prior works \cite{epstein2020oops,benaim2020speednet,yao2020video,jenni2020video}.\\

\noindent\textbf{Direction Classification.} 
In this case, the temporal transformations can result in videos being played backward.
As in the speed classification, this transformation is predictable from a single transformed example. 
Thus, to predict this transformation, we train a binary classifier on top of the learned video representation $F(x)$.
This self-supervised task was proposed in \cite{wei2018learning}.\\

\noindent\textbf{Overlap-/Order Classification}.
Finally, we propose a self-supervised classification task concerning relative time-shifts between two clips, \ie, this task concerns a pair $(\tau_p(x), \tau_q(x))$ with $\tau_i$ containing temporal shifts. 
We discretize the event space into three distinct classes: 1. $\tau_p(x)$ comes entirely before $\tau_q(x)$, 2. $\tau_p(x)$ comes entirely after $\tau_q(x)$, and 3. $\tau_p(x)$ and $\tau_q(x)$ temporally overlap.
We solve this task by feeding the concatenated feature vectors, \ie, $[F(\tau_p(x))^\top, F(\tau_q(x))^\top]$ to a non-linear classifier.\\

Note that accurately solving the temporal equivariance task in Eq.\ref{eq:equi} subsumes all these auxiliary SSL tasks. 
Figure \ref{fig:transforms} illustrates the temporal transformations we consider, the different auxiliary tasks, and how the equivariance objective relates to them.

\begin{figure}
    \centering
    \includegraphics[width=\linewidth]{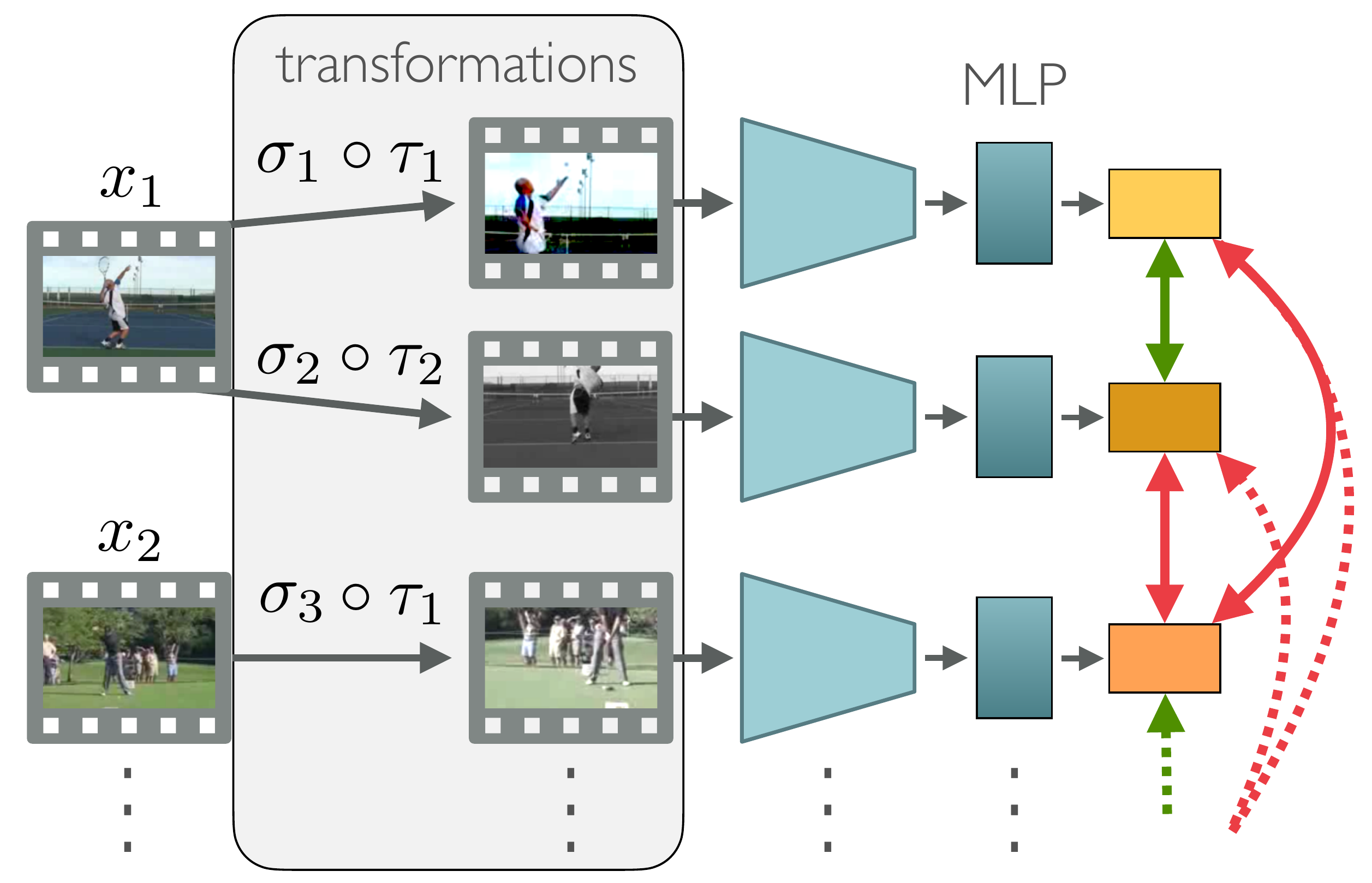}
    \caption{\textbf{Instance discrimination task.} We illustrate the task of instance discrimination which we perform on the same batch of examples used for the equivariance objective (compare to Fig.\ref{fig:equi_model}). Augmentations of the same example are attracted, and other examples are repelled. Note that the MLPs in the two cases are different.   }
    \label{fig:inst_model}
\end{figure}

\subsection{Instance Discrimination Objective}\label{sec:inst}

Besides learning equivariance to certain input transformations $\mathcal{T}$, we also want the learned video representation $F(x)$ to be sensitive to the actual input scene changing.
We achieve this by adding a standard instance contrastive learning objective \cite{chen2020simple}. 
Let $\Tilde{x}_i^p=\sigma_p \circ \tau_p(x_i)$ denote an augmented training example with some randomly sampled $\sigma_p\sim \mathcal{S}$ and $\tau_p\sim \mathcal{T}$.
To obtain instance discriminative features, we encode these augmented examples in a vector $\phi_i^p=\phi(F(\Tilde{x}_i^p))\in \mathbb{R}^D$, where $\phi$ represents a multi-layer perceptron.
We can now formualte the instance discrimination objective as
\begin{equation}
    \mathcal{L}_{\operatorname{inst}}=-\mathbb{E}\left[\log \frac{d(\phi_i^p , \phi_i^q)}{d(\phi_i^p , \phi_i^q) +\sum_{j\neq i} d(\phi_i^p , \phi_j^r)}\right],
    \label{eq:inst}
\end{equation}
where $d(\cdot, \cdot)$ is again defined as in Eq.\ref{eq:sim}.
The training on the instance discrimination objective is illustrated in Figure \ref{fig:inst_model}.
As shown in the Figure, we perform the two objectives in Equation \ref{eq:equi} and \ref{eq:inst} on the same examples in the mini-batch. 
However, the two MLPs $\psi$ and $\phi$ learn different non-linear projections of $F(x)$ that capture changes in the input transformations and input instance, respectively.



\subsection{Implementation Details}
Networks were trained using the AdamW optimizer \cite{loshchilov2018decoupled} with default parameters and a weight decay of $10^{-4}$. 
During pre-training, the learning rate is first linearly increased to $3\cdot10^{-4}$ ($10^{-4}$ for fine-tuning) and then decayed with a cosine annealing schedule \cite{loshchilov2016sgdr}.
In addition to the temporal input transformations, we also use standard data-augmentations used in contrastive methods, \eg, horizontal flipping, color-jittering, and random cropping. 
We use a set of different backbone architectures (3D-ResNets\cite{hara2018can}, R(2+1)D \cite{tran2018closer}, S3D-G \cite{xie2018rethinking}) and always consider as learned representation $F(x)$ the output of the global pooling layer. 
The MLPs $\phi(\cdot)$ and $\psi(\cdot)$ each have two hidden layers and preserve the original feature dimension.
The various classification heads for the auxiliary SSL tasks each have a single hidden layer.
Input clips contain 16 frames at a resolution of $128\times 128$ if not specified otherwise, and we skip three frames during transfer learning (corresponding to the $4\times$ speed encountered during pre-training).  
We perform inference by averaging predictions of multiple temporal and spatial crops following prior works \cite{han2020self,han2020memory}.
Features for linear probes and nearest-neighbor retrieval are similarly obtained by averaging features from multiple crops and standardizing the result based on training set statistics.


\begin{table*}[ht]
\centering
\caption{\textbf{Ablation experiments.}
We illustrate the influence of different input equivariances (a)-(d), the different self-supervised learning objectives (e)-(k), and the composition of the auxiliary SSL objective (l)-(o) on the feature performance.  
We report action recognition accuracy using linear classifiers, full fine-tuning, and nearest-neighbor classification.
}\label{tab:ablations}
\begin{tabular}{@{}l@{\hspace{.5em}}l@{\hspace{1em}}c@{\hspace{1em}}c@{\hspace{1em}}c@{\hspace{2em}}c@{\hspace{1em}}c@{\hspace{1em}}c@{}}
\toprule
 & & \multicolumn{3}{c}{\textbf{UCF101}} & \multicolumn{3}{c}{\textbf{HMDB51}}  \\ 
\multicolumn{2}{l}{\textbf{Ablation}} & Linear & Finetune  & 1-NN & Linear & Finetune  & 1-NN \\ \midrule
(a) & no equivariance & 54.2 & 73.4 & 48.7 & 25.2 & 45.4 & 17.8  \\  
(b) & spatial only    &  62.8 &  75.5  &  45.6  & 38.2  &  48.9 & 19.3   \\
(c) & spatial + temporal    & \underline{70.7}  &  \underline{80.8} &  \underline{53.0}  & \underline{44.1}  &  \underline{57.6} &  \underline{27.5}  \\
(d) & temporal only  & \textbf{74.1} & \textbf{83.7} & \textbf{62.1} & \textbf{47.5} & \textbf{60.8} & \textbf{31.5}  \\
\midrule
(e) & $\mathcal{L}_{inst}$ & 54.2 & 73.4 & 48.7 & 25.2 & 45.4 & 17.8  \\  
(f) & $\mathcal{L}_{equi}$ & 36.2 & 74.5 & 21.2 & 21.2 & 52.2 & 10.7  \\
(g) & $\mathcal{L}_{aux}$ & 67.5 & \textbf{84.2} & 49.8 & 41.7 & 59.6 & 26.6  \\
(h) & $\mathcal{L}_{inst}+\mathcal{L}_{equi}$ & 63.4 & 77.0 & 52.1 & 37.4 & 48.8 & 21.4  \\ 
(i) & $\mathcal{L}_{inst}+\mathcal{L}_{aux}$ & \underline{71.8} & 83.0 & \underline{59.9} & \underline{44.9} & 58.8 & \underline{30.6}  \\
(j) & $\mathcal{L}_{equi}+\mathcal{L}_{aux}$ & 67.1 & 83.3 & 48.0 & 42.8 & \textbf{61.0} & 27.0  \\
(k) & $\mathcal{L}_{inst}+\mathcal{L}_{equi}+\mathcal{L}_{aux}$ & \textbf{74.1} & \underline{83.7} & \textbf{62.1} & \textbf{47.5} & \underline{60.8} & \textbf{31.5}  \\
\midrule
(l) & $\texttt{aux}=\texttt{speed}$ & 69.0 & 82.3 & 56.6 & 42.0 & 58.3 & 27.4  \\  
(m) & $\texttt{aux}=\texttt{speed}+\texttt{order}$ & 70.1 & 81.6 & 57.9 & 44.3 & 58.6 & \underline{29.0}  \\  
(n) & $\texttt{aux}=\texttt{speed}+\texttt{rev}$ & \underline{71.1} & \underline{83.4} & \underline{60.0} & \underline{45.6} & \underline{59.7} & 27.5  \\  
(o) & $\texttt{aux}=\texttt{speed}+\texttt{rev}+\texttt{order}$ & \textbf{74.1} & \textbf{83.7} & \textbf{62.1} & \textbf{47.5} & \textbf{60.8} & \textbf{31.5}  \\
\bottomrule 
\end{tabular}
\label{tab.ablations}
\end{table*}

\section{Experiments}

\noindent\textbf{Datasets.} We consider four datasets in our experiments. Kinetics-400 \cite{zisserman2017kinetics} contains around 240K training videos and is used for unsupervised pre-training, \ie, we do not use the action labels. 
UCF101 \cite{soomro2012ucf101} and HMDB51 \cite{hmdb51} are smaller datasets with human action labels and are used to evaluate the learned representation in transfer to action recognition via fine-tuning and as fixed feature extractors for video retrieval. 
We also use UCF101 training split 1 for unsupervised pre-training in ablation and retrieval experiments. 
Finally, we evaluate our method on Diving48 \cite{li2018resound}, a dataset in which different action classes differ primarily in their long-range motion patterns rather than static frame appearance.

\begin{table*}[]
    \centering
    \caption{\textbf{Comparison to prior work on self-supervised video representation learning.} We report action recognition accuracy after fine-tuning to UCF101 and HMDB51. We indicate the pre-training dataset, input resolution, number of input frames, network architecture, and pre-training data modality (V=RGB, F=optical-flow, A=audio, T=text).  }
    \label{tab:comparison}
    \begin{tabular}{@{}l@{\hspace{1em}}c@{\hspace{1em}}c@{\hspace{1em}}c@{\hspace{1em}}c@{\hspace{1em}}c@{\hspace{1em}}c@{\hspace{1em}}c}
    \toprule
    \textbf{Method}      &  \textbf{Dataset} & \textbf{Res.}  &   \textbf{Frames} &    \textbf{Network}  &   \textbf{Mod.}    &   \textbf{UCF101}    &   \textbf{HMDB51} \\ \midrule
    
    VCOP \cite{xu2019self} & UCF101  & 112  &  16   &  R(2+1)D    &    V  &  {72.4}   &   30.9  \\
    PRP \cite{yao2020video} & UCF101  &   112   & 16  &  R(2+1)D    &   V    &  72.1   &   {35.0}  \\
    Var. PSP \cite{cho2020self}  & UCF101 &   112   & 16  &  R(2+1)D   &   V   &  {74.8}   & {36.8}   \\
    Temp.-Trans. \cite{jenni2020video}  & UCF101 &   112   & 16  &  R(2+1)D   &   V   &  {81.6}   & {46.4}   \\
    Pace Pred. \cite{wang2020self} & Kinetics-400  & 112  &  16   &  R(2+1)D    &    V  &  77.1   &   36.6  \\
    VideoDIM \cite{devon2020representation} & Kinetics-400  & 128  &  32   &  R(2+1)D    &    V  &  79.7   &   49.2  \\
    TCLR \cite{dave2021tclr} & Kinetics-400  &  112  & 16  & R(2+1)D    &  V    &  84.3   &   54.2  \\
    CBT \cite{sun2019contrastive}   &   Kinetics-600  &   112   & 16  &   S3D  &   V   &   79.5  & 44.6   \\
    SpeedNet \cite{benaim2020speednet}  &  Kinetics-400 &  224 & 64 &  S3D-G  & V  &   81.1  & 48.8   \\
    3D ST-puzzle \cite{kim2019self} & Kinetics-400  &  224  & 16  & R3D-18    &  V    &  65.8   &   33.7  \\
    MemDPC \cite{han2020memory}   &  Kinetics-400 &  224 &  40  &  R3D-34   &   V   &   78.1  & 41.2   \\
    CVRL \cite{qian2020spatiotemporal} & Kinetics-400  &  224  & 32  & R3D-50    &  V    &  92.1   &   65.4  \\
    \midrule
    STS \cite{wang2020self2} & Kinetics-400  &  224  & 64  &  S3D-G    &  V+F    &  89.0   &   62.0  \\
    CoCLR \cite{han2020self} & Kinetics-400  &  128  & 32  & S3D    &  V+F    &  87.9   &   54.6  \\
    AVTS \cite{korbar2018cooperative} &  Kinetics-400 &  224   &  25  &   MC3 &  V+A   &   85.8  & 56.9   \\
    XDC \cite{alwassel2019self} &  Kinetics-400 &  224   &  8  &   R(2+1)D &  V+A   &   84.2  & 47.1   \\
    GDT \cite{patrick2020multi} &  Kinetics-400 &  112   &  32  &   R(2+1)D &  V+A   &   89.3  &  60.0   \\
    MIL-NCE \cite{miech2020end} &  HowTo100M &  224   &  32  &   S3D &  V+T   &   91.3  &  61.0   \\
    \midrule
    \textbf{Ours} &  Kinetics-400 &  128    &  16  &   R3D-18 &  V   &    87.1  &   63.6   \\
    \textbf{Ours} &  Kinetics-400 &  112    &  16  &   R(2+1)D &  V   &    88.2  &   62.2   \\
    \textbf{Ours} &  Kinetics-400 &  128    &  32  &   S3D-G &  V   &    86.9  &   63.5   \\
    \bottomrule
    \end{tabular}
\end{table*}

\subsection{Ablation Experiments}

\begin{figure}
    \centering
    \includegraphics[width=\linewidth]{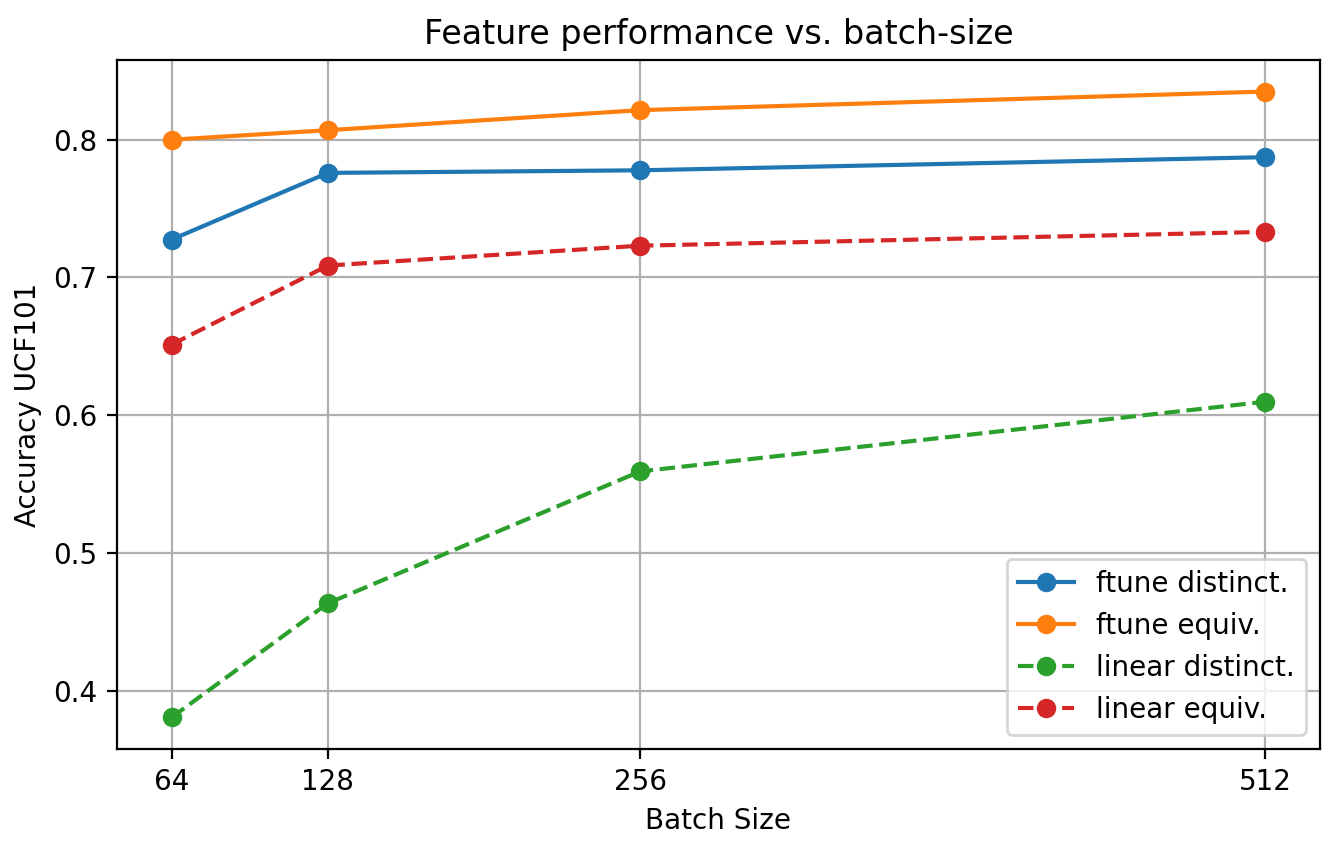}
    \caption{\textbf{Equivariance vs. distinctiveness.} We show how contrastive equivariance learning scales with the number of negative examples (\ie, batch size) and compare to a model merely contrasting between temporal augmentations (\ie, learning distinctiveness).
    }
    \label{fig:batch_plot}
\end{figure}

We perform ablation experiments to investigate the effect of learning equivariance to different types of input transformations and the influence of the various self-supervised learning objectives. 
We pre-train a 3D-ResNet-18 network for 400 epochs on UCF101 training split 1 with a batch size of 192. 
The learned representations are then evaluated on UCF101 and HMDB51 via linear SVM classifiers, full finetuning, and nearest neighbors on the action recognition task.
Results of the following ablations are summarized in Table \ref{tab:ablations}:\\

\noindent \textbf{(a)-(d) Equivariance vs. invariance:} We compare models trained without equivariance (a), equivariance to only spatial transformations (b), spatial and temporal equivariance (c), and only temporal equivariance (d).  
These experiments thus change the composition of $\mathcal{T}$ and $\mathcal{S}$, \eg, in (b) $\mathcal{T}$ consists of only spatial augmentations. 
The model with temporal equivariance and spatial invariance performs best in all metrics. Interestingly, we observe that spatial equivariance leads to better performance in most cases compared to models without any equivariance training. Note that we used consistent temporal augmentations for spatial equivariance learning to remove possible ambiguities. 

\noindent \textbf{(e)-(k) Training objectives:} We investigate how feature performance is influenced by the different training objectives concerning temporal equivariance $\mathcal{L}_{equi}$, instance discrimination $\mathcal{L}_{inst}$, and auxiliary SLL tasks $\mathcal{L}_{aux}$. Overall, we observe that the combination of all the objectives performs best, especially regarding fixed-feature evaluation via linear and nearest neighbor classification.
Interestingly, the cases (g) and (j) training only for temporal equivariance perform best in finetuning, suggesting that representations with a strong initial bias towards motion generalize better with few labeled examples. 
Noteworthy is also the importance of $\mathcal{L}_{aux}$ in combination with $\mathcal{L}_{equi}$. Indeed, we find that networks are unable to optimize $\mathcal{L}_{equi}$ well without the guiding training signals from  $\mathcal{L}_{aux}$.

\noindent \textbf{(e)-(k) Auxillary SSL tasks:} We demonstrate the effect of the different auxiliary SSL tasks, \ie, speed prediction, direction prediction, and the proposed clip-order/overlap prediction.  Each of these tasks corresponds to different temporal transformation types (speed changes, playback direction, and temporal shifts), all of which are beneficial and improve the equivariance objective's optimization.\\

Finally, in Figure \ref{fig:batch_plot} we show how our models' performance scales with the number of negative samples available for contrastive learning. 
We compare this to a model that is trained to be distinctive, but not equivariant, to temporal transformations.
Besides superior performance, we observe that the equivariance model achieves good performance even with small batch sizes.
This is of practical value given the large memory footprint of 3D-CNNs.

\subsection{Comparison to Prior Works}

\begin{table*}[h]
    \centering
    \caption{\textbf{Video Retrieval Performance on UCF101 and HMDB51.} We report recall at $k$ (R@$k$) for $k$-NN based video retrieval. Query videos are taken from test split 1 and retrievals computed on train split 1 of UCF101 and HMDB, respectively. * indicates Kinetics pre-training.  }
    \label{tab:nn}
    \begin{tabular}{@{}l@{\hspace{1em}}c@{\hspace{1em}}c@{\hspace{1em}}c@{\hspace{1em}}c@{\hspace{1em}}c@{\hspace{2em}}c@{\hspace{1em}}c@{\hspace{1em}}c@{\hspace{1em}}c}
    \toprule
         &   &  \multicolumn{4}{c}{\textbf{UCF101}}  &  \multicolumn{4}{c}{\textbf{HMDB51}}  \\ 

    \textbf{Method}      & \textbf{Network} &  \textbf{R@1}  &  \textbf{R@5} &   \textbf{R@10}   &   \textbf{R@20}  &   \textbf{R@1}  &  \textbf{R@5} &   \textbf{R@10}   &   \textbf{R@20}    \\ \midrule
    OPN \cite{lee2017unsupervised} &  AlexNet  &   19.9 & 28.7 & 34.0 & 40.6 
    & - & - & - & -\\     
    B\"uchler \etal~\cite{buchler2018improving} &  AlexNet  &   {25.7} & 36.2 & 42.2 &  49.2 & - & - & - & - \\

    STS \cite{wang2020self2} &  C3D    &  30.1  & 49.6  & 58.8 & 67.6
    & 13.9 & 33.3 & 44.7 & 59.5 \\   
    Pace Pred. \cite{wang2020self} &  C3D    &  31.9  & 49.7  & 59.2 & 68.9
    & 12.5 & 32.2 & 45.4 & 61.0 \\ 
    PRP \cite{yao2020video} &  R3D-18    &  22.8  & {38.5}  & {46.7}  & {55.2}
    & - & - & - & - \\
  
    VCOP \cite{xu2019self} &  R3D-18    &  14.1  & 30.3  & 40.4 & 51.1
    & 7.6 & 22.9 & 34.4 & 48.0 \\
    VCP \cite{luo2020video} &  R3D-18    &  18.6  & 33.6  & 42.5 & 53.5
    & 7.6 & 24.4 & 36.6 & 53.6 \\
    
    Var. PSP \cite{cho2020self}  &  R3D-18    &  24.6  & 41.9  & 51.3 & 62.7
    & 10.3 & 26.6 & 38.8 & 51.6 \\
    
    PCL \cite{tao2020self}  &  R3D-18    &   40.5  &  59.4  & 68.9  &  77.4
    &  16.8 &  38.4 &  53.4 & 68.9  \\
                 
    MemDPC \cite{han2020memory}  &  R3D-18    &  20.2  & 40.4  & 52.4 & 64.7
    & 7.7 & 25.7 & 40.6 & 57.7 \\
    Temp.-Trans. \cite{jenni2020video}* &  R3D-18  &  {26.1} & {48.5} & {59.1} & {69.6} 
    & - & - & - & - \\  
    SpeedNet \cite{benaim2020speednet}*   &  S3D-G  &  13.0 & 28.1 & 37.5 & 49.5 
    & - & - & - & -\\
    CoCLR \cite{han2020self} &  S3D    &  53.3  & 69.4  & 76.6 & 82.0
    & 23.2 & 43.2 & 53.5 & 65.5 \\

    TCLR \cite{dave2021tclr} &  R(2+1)D    & {56.9}  & {72.2}  & {79.0} & {84.6}
    & {24.1} & {45.8} & {58.3} & \underline{75.3} \\

    GDT \cite{patrick2020multi}* &  R(2+1)D    & \underline{57.4}  & \underline{73.4}  & \underline{80.8} & \underline{88.1}
    & \underline{25.4} & \underline{51.4} & \underline{63.9} & {75.0} \\
    \midrule
    \textbf{Ours} &  R3D-18    &  {63.6}  & {79.0}  & {84.8} & {89.9}
    & \textbf{32.2} & \textbf{60.3} & \textbf{71.6} & \textbf{81.5} \\
    \textbf{Ours} &  R(2+1)D    &  \textbf{64.3}  & \textbf{80.9}  & \textbf{86.4} & \textbf{90.6}
    & {29.5} & {55.8} & {68.0} & {78.2} \\
    
    \bottomrule
    \end{tabular}
\end{table*}

\noindent \textbf{Action recognition on UCF101 and HMDB51.}
The most common evaluation of self-supervised video representations is via complete finetuning for action recognition on UCF101 and HMDB51. 
We compare to prior works using three common backbone architectures. 
Networks are pre-trained on Kinetics-400 for 200 epochs.
The batch size is set to 512 for 3D-ResNet-18, 256 for S3D-G, and 192 for R(2+1)D (adjusting for those networks' different memory footprints).
Finetuning is performed for 100 epochs on UCF101 and 200 epochs on HMDB51 using a batch-size of 32.
Our results and a comparison to prior works can be found in Table \ref{tab:comparison}.
A fair comparison to prior works is difficult since network architectures, pre-training datasets, data modalities, and evaluation protocols can vary significantly. 
We indicate input resolution, the number of frames, network architecture, and pre-training data modalities, all of which influence performance, in the table. 
When possible, we report numbers that correspond most closely to our setting, \ie, we report results using only RGB videos at test time when available. 

Our method achieves very competitive performance - especially on HMDB51 - with all tested architectures and outperforms other RGB-only approaches, except for \cite{qian2020spatiotemporal} which is based on a considerably larger architecture with much higher computational requirements.

\noindent \textbf{Video Retrieval on UCF101 and HMDB51.}
We also compare to prior works on the video retrieval task in Table \ref{tab:nn}. 
Nearest-neighbors are computed using cosine similarity, and we use networks pre-trained on UCF101 in this evaluation, following the majority of prior work.
We observe strong performance with our method and achieve state-of-the-art results on both datasets. 
This is consistent with the drastic improvements resulting from combining temporal equivariance and instance discrimination learning, as observed in the ablations (see Table \ref{tab:ablations}). 
The results also show the benefit of learning temporal equivariance compared to models that merely learn to distinguish between temporal transformations \cite{dave2021tclr,patrick2020multi}.  


\begin{table}[t]
    \centering
    \caption{\textbf{Comparison on Diving48.} We report classification accuracy using old (\textbf{V1}) and updated (\textbf{V2}) diving labels. We indicate playback speed for our results. Prior works all use $1\times$ speed.  }

    \begin{tabular}{@{}l@{\hspace{2em}}c@{\hspace{1em}}c@{}}
    \toprule
      &   \multicolumn{2}{c}{\textbf{Accuracy}}   \\ 

    \textbf{Method}    & \textbf{V1}  &   \textbf{V2} \\
    \midrule 

    Random Init. ($1\times$)      &  18.8   & 50.7  \\ 
    Random Init. ($2\times$)     &  26.6   & 64.3  \\ 
    \midrule
    RESOUND-C3D \cite{li2018resound}     &  16.4   &   -  \\ 
    TSN \cite{wang2016temporal}      &  16.8   &   -  \\ 
    Debiased R3D-18 \cite{choi2019can}    &  20.5   &   -  \\ 

    TCRL \cite{dave2021tclr}    &  22.9   &   -  \\ 

    \midrule
    \textbf{Ours} ($1\times$)    &  29.9   &  71.2 \\
    \textbf{Ours}  ($2\times$)    &  34.9   &  76.2 \\

    \bottomrule
    \end{tabular}
    \label{tab:my_label}
\end{table}

\noindent \textbf{Evaluation on Diving48.} Finally, we evaluate our method on Diving48 \cite{li2018resound}, which is a challenging dataset as it requires capturing long-term motion patterns to recognize the classes accurately. 
Furthermore, unlike in other benchmarks, appearance does not correlate strongly with the label.
We pre-train an R(2+1)D on the Diving48 training set and follow the most common evaluation using 16 consecutive input frames (which corresponds to $1\times$ playback in our method). 
For completeness, we also report results using $2\times$ playback (\ie, covering 32 input frames) and results on updated action labels (\textbf{V2}).
The strong inductive motion bias in our method leads to state-of-the-art results.


\section{Conclusions}
We introduced a novel self-supervised learning approach to learn video representations that are equivariant to temporal input transformations. 
This method is motivated by the importance of dynamics for video understanding and the desire to reflect variations in the input motion patterns. 
Our method extends the contrastive learning framework with equivariance constraints on relative temporal transformations between augmented samples. 
Experiments demonstrate that representations with temporal equivariance achieve state-of-the-art performance on classic vision tasks such as action recognition or video retrieval.

{\small
\bibliographystyle{ieee_fullname}
\bibliography{egbib}
}

\end{document}